%% file: acl_latex.tex
\pdfoutput=1

\documentclass[11pt]{article}

\usepackage[final]{acl}
\usepackage{multirow}
\usepackage{times}
\usepackage{latexsym}
\usepackage{booktabs}
\usepackage{caption}
\usepackage[T1]{fontenc}

\usepackage[utf8]{inputenc}

\usepackage{microtype}

\usepackage{inconsolata}
\usepackage{tikz}
\def\checkmark{\tikz\fill[scale=0.4](0,.35) -- (.25,0) -- (1,.7) -- (.25,.15) -- cycle;} 
\usepackage{graphicx}
\usepackage{tabularx}

\usepackage{color-edits}
\addauthor[Alexis]{ap}{cyan}

\title{Untangling the Influence of Typology, Data and Model Architecture on Ranking Transfer Languages for Cross-Lingual POS Tagging}

\author{Enora Rice${ }^{1}$ \quad Ali Marashian${ }^{1}$ \quad  Hannah Haynie${ }^{1}$  \quad 
        Katharina von der Wense${}^{1,2}$ \\ {\bf Alexis Palmer}${ }^{1}$ \\
    ${ }^{1}$University of Colorado Boulder \quad ${ }^{2}$ Johannes Gutenberg University Mainz  \\ \texttt{enora.rice@colorado.edu}}

\begin{document}
\maketitle
\begin{abstract}

Cross-lingual transfer learning is an invaluable tool for overcoming data scarcity, yet selecting a suitable transfer language remains a challenge. The precise roles of linguistic typology, training data, and model architecture in transfer language choice are not fully understood. We take a holistic approach, examining how both dataset-specific and fine-grained typological features influence transfer language selection for part-of-speech tagging, considering two different sources for morphosyntactic features. While previous work examines these dynamics in the context of bilingual biLSTMS, we extend our analysis to a more modern transfer learning pipeline: zero-shot prediction with \textbf{pretrained multilingual models}. We train a series of transfer language ranking systems and examine how different feature inputs influence ranker performance across architectures. Word overlap, type-token ratio, and genealogical distance emerge as top features across all architectures. Our findings reveal that a combination of \textbf{typological} and \textbf{dataset-dependent} features leads to the best rankings, and that good performance can be obtained with either feature group on its own.

\end{abstract}

\section{Introduction}
\input{alt_intro}

\section{Related Works}

\subsection{Ranking Transfer Languages}
\citet{lin-etal-2019-choosing} rank transfer languages using both dataset-dependent and linguistic features from the URIEL knowledge base \cite{littell-etal-2017-uriel}.  We build on their work with key adaptations: 1) Instead of varying dataset size, which obscures the role of fine-grained features, we hold corpus size constant across all language pairs.
2) 
In addition to bilingual biLSTMs, we examine zero-shot transfer with finetuned MLMs.
3) We replace typological distance measures with element-wise comparisons of typological feature vectors, following \citet{DBLP:journals/corr/abs-2105-05975}. 

\citet{khan-etal-2025-uriel} build on the work in \citet{littell-etal-2017-uriel} to enhance the coverage of URIEL and lang2vec with novel linguistic databases and customizable distance calculations. We follow suit by comparing the impact of incorporating URIEL syntactic vectors versus Grambank syntactic vectors on the transfer language ranking task

\subsection{Transfer Language Choice for Zero-shot Cross-lingual Transfer with MLMs}
\citet{Lauscher_Ravishankar_Vulić_Glavaš_2020} show a correlation between linguistic proximity and successful zero-shot transfer, but only test English as the source language. We experiment with 18 source languages. \citet{de-vries-etal-2022-make} find that XLM-R finetuned on a suitable transfer language performs almost three times better than when using a suboptimal transfer language. They highlight the influence of linguistic similarity but do not consider dataset features. 

\section{Experiments}
\subsection{Languages}
We experiment with a total of 20 target and 18 source languages across seven language families. We determine our set of target and source languages based on the availability of sufficient data in Universal Dependencies 2.0 (UD) \citep{de_Marneffe_Manning_Nivre_Zeman_2021}. We consider target languages that have a training corpus with at least 500 lines and source languages with at least 2000. Justification for this threshold is described in \ref{sec:bi-lstms}. We also eliminate languages that are not present in URIEL and/or Grambank. Our full set of target languages is given in Table \ref{tab:langs}. Languages that also serve as source languages are italicized. While many of the languages covered by our experiments are high-resource, several others fall into a middle range and are undeserved by the NLP research community at large.

\subsection{Testbed Tasks}
\label{sec:testbed}
We generate gold ranking-data by training a suite of biLSTMs and finetuned XLM-R and M-BERT models for POS tagging across all possible source/target language pairs. To remove the influence of dataset size, we cap each source language training set at 2000 lines. Then, for each target language, we create a ranking of all potential source languages based on the relative performance of each model on a held out test set. Model details are outlined in following sections. 

\subsubsection{biLSTMs}
\label{sec:bi-lstms}
We train a suite of 378 biLSTMs using Stanza \citep{qi-etal-2020-stanza}--  one for each target/source pair. We train each model on 500 instances of UD data in the target language and 2000 instances in the source language. We choose this split to simulate a setting where limited training data is available in the target language but comparatively greater data is available in the source language. We set the data thresholds to ensure that sufficient training data is present for model convergence, but training data in the target language is still limited enough to make the task non-trivial. All models are trained on default Stanza hyperparameters \textit{without} pre-trained word embeddings for a maximum of 6000 steps. We evaluate each model on a held out test set drawn from the same corpus as the target training data.

\subsubsection{Fine-tuned XLM-R and M-BERT}
We finetune XLM-R and M-BERT equivalently on each of our 18 source languages with a modified implementation\footnote{https://github.com/wietsedv/xpos} from 
\citet{de-vries-etal-2022-make}.
Each model is trained on the same 2000 instance UD dataset that we use to train our biLSTM models. All models are trained for 1,000 batches of 10 samples with a linearly decreasing learning rate starting at 5e-5. We use 10\% dropout between transformer layers and 10\% self-attention dropout.

\begin{table}[h!]
\centering
\begin{tabular}{|l|l|}
\hline
\textbf{Language} & \textbf{Treebank} \\ \hline
\textit{Basque}& UD\_Basque-BDT         \\ \hline
\textit{Czech}& UD\_Czech-PDT          \\ \hline
\textit{Danish}& UD\_Danish-DDT         \\ \hline
\textit{Dutch}& UD\_Dutch-LassySmall   \\ \hline
Finnish   & UD\_Finnish-FTB        \\ \hline
\textit{Hindi}& UD\_Hindi-HDTB         \\ \hline
Hungarian & UD\_Hungarian-Szeged   \\ \hline
\textit{Indonesian}& UD\_Indonesian-GSD    \\ \hline
Galician  & UD\_Galician-CTG       \\ \hline
\textit{Italian}& UD\_Italian-PoSTWITA   \\ \hline
\textit{Korean}& UD\_Korean-GSD         \\ \hline
\textit{Latin}& UD\_Latin-ITTB         \\ \hline
\textit{Latvian}& UD\_Latvian-LVTB       \\ \hline
\textit{Turkish}& UD\_Turkish-IMST       \\ \hline
\textit{Polish}& UD\_Polish-LFG         \\ \hline
\textit{Portuguese}& UD\_Portuguese-Bosque \\ \hline
\textit{Russian}& UD\_Russian-SynTagRus  \\ \hline
\textit{Catalan}& UD\_Catalan-AnCora     \\ \hline
\textit{French}& UD\_French-Sequoia     \\ \hline
\textit{English}& UD\_English-LinES      \\ \hline
\textit{Ukrainian}& UD\_Ukrainian-IU       \\ \hline
\end{tabular}
\caption{Full list of target languages and their corresponding treebanks. Languages that also serve as source languages are italicized.}
\label{tab:langs}
\end{table}

\subsection{Our Ranking System} 
Given a target language $t$ and a list of $n$ potential source languages $S = [s_1,s_2...s_n]$, our goal is to rank all source languages in $S$ based on the expected performance of POS-tagging models trained on each source/target pair ($s_i$,$t$). 
Building on \citet{lin-etal-2019-choosing}, we train a series of gradient boosted decision trees using the LightGBM implementation (MIT License) \citep{Ke2017LightGBMAH} of the LambdaRank algorithm. Models are trained on  gold ranking-data described in Section \ref{sec:testbed}. 

Input to our ranking system consists of vector representations of each source/target pair. Vectors are defined as a set of features, categorized into two types. We calculate \textbf{dataset-dependent} features by comparing source and target corpora using four metrics: word overlap, type-token ratio in the source language corpus, type-token ratio in the target language corpus, and the difference between the source and target language type-token ratios. \textbf{Dataset-independent} features capture linguistic similarity between the source and target languages using five measures: \textit{genetic}, \textit{syntactic}, \textit{phonological}, (phonetic) \textit{inventory}, and \textit{geographic}. \textit{Syntactic}, \textit{phonological} and \textit{inventory} features are defined using binary feature vectors sourced from typological databases. We call these our \textit{Typology-Vector} features. By default, Typology-Vector features are represented by distance measures computed as the cosine difference between URIEL \citep{littell-etal-2017-uriel} vectors representing source and target, but we experiment with different representations (described in Sections \ref{sec:dist} and \ref{sec:grambank}). All features are briefly summarized in Table \ref{tab:features} and feature vector lengths are given in Table \ref{tab:feature_lengths}. For more detailed descriptions, refer to \citet{lin-etal-2019-choosing}.

\begin{table*}[h!]
\centering
\small
\begin{tabular}{|p{0.30\linewidth}|p{0.60\linewidth}|}
\hline
\textbf{Feature Type}& \textbf{Description} \\ \hline
\textit{Genetic} Distance& Genealogical distance derived from language descent trees described in Glottolog.\\ \hline
\textit{Geographic} Distance& Defined as the orthodromic distance divided by the antipodal distance between rough locations of source and target languages on the surface of the Earth. \\ \hline \textit{Syntactic}, \textit{Phonological}, and  \textit{Inventory} Distances (\textbf{distance} Typology-Vector)&  Computed as the cosine difference between  corresponding URIEL \citep{littell-etal-2017-uriel} or Grambank \citep{Skirgård_Haynie_Blasi_Hammarström_Collins_Latarche_Lesage_Weber_Witzlack-Makarevich_Passmore_et_al._2023} feature vectors representing source and target languages.\\
 \textit{Syntactic}, \textit{Phonological}, and  \textit{Inventory} Vectors (\textbf{full} Typology-Vector)&Computed as element-wise AND operation between  corresponding URIEL  \citep{littell-etal-2017-uriel} or Grambank \citep{Skirgård_Haynie_Blasi_Hammarström_Collins_Latarche_Lesage_Weber_Witzlack-Makarevich_Passmore_et_al._2023} feature vectors representing source and target languages.\\ \hline Dataset-Dependent Features & Word overlap, transfer type-token ration, source type-token ration, type-token ratio distance \\\hline
\end{tabular}
\caption{All possible ranker features}
\label{tab:features}
\end{table*}

\begin{table}[h!]
\centering
\small
\begin{tabular}{|l|c|}
\hline
\textbf{Vector Type}& \textbf{Description} \\ \hline
\textit{URIEL Syntactic} & 104\\ \hline
\textit{Grambank Syntactic}& 113\\ \hline \textit{Phonological}&  28\\ \hline
 \textit{Inventory}&158\\ \hline
\end{tabular}
\caption{Typological feature vector lengths}
\label{tab:feature_lengths}
\end{table}

\subsubsection{Distance-Measure vs. Fully Featured}
\label{sec:dist}
By default, we express the linguistic similarity between \textit{syntactic}, \textit{phonological}, and \textit{inventory} features as a series of distance measures. We call these \textbf{distance} Typology-Vector representations. At predict time, the ranker receives a feature vector $a$ representing the target and a feature vector $b$ representing the source and computes the cosine distance: $1 - cos(a,b) = d$.
We concatenate $d$ to the final ranking model input vector.

To analyze the impact of fine-grained features on transfer language suitability, we experiment with an expanded representation, using an element-wise \textit{and} operation to compare $a$ and $b$: $a\land b = v$. We refer to $v$ as the \textbf{full} Typology-Vector representation. We concatenate $v$ to ranker input.

\subsubsection{URIEL vs. Grambank}
\label{sec:grambank}
Many typological analyses of crosslingual transfer rely on URIEL (CC BY-SA 4.0) feature vectors, which are heavily based on the World Atlas of Language Structures (CC BY 4.0) \citep{wals}.  WALS has incomplete genealogical coverage and over 80\% missing data \citep{Skirgård_Haynie_Blasi_Hammarström_Collins_Latarche_Lesage_Weber_Witzlack-Makarevich_Passmore_et_al._2023}. As such, we experiment with switching to Grambank (CC BY 4.0) \citep{Skirgård_Haynie_Blasi_Hammarström_Collins_Latarche_Lesage_Weber_Witzlack-Makarevich_Passmore_et_al._2023}, which addresses some of WALS' shortcomings. We impute all undefined features in either database as follows.
\paragraph{URIEL.} We use URIEL vectors that have been pre-imputed by \citet{littell-etal-2017-uriel} using k-nearest-neighbors.\footnote{vectors available at https://github.com/antonisa/lang2vec}

\paragraph{Grambank.} 24\% of  total feature values in Grambank 1.0.3 (across all languages in the database) are undefined. In order to produce fully defined feature vectors for our experiments, we first eliminate any features that are undefined for greater than 25\% of languages and any languages that have greater than 25\% missing data. After cropping, only 4.03\% of values are missing. We impute the remaining values with the MissForest algorithm for nonparametric missing value imputation \citep{Stekhoven_Bühlmann_2012}. We adapt our imputation procedure from \citet{grambank_dataset_zenodo_v1}.

\begin{table*}[!t]
\setlength{\tabcolsep}{1pt}
    \small
    \centering
    \begin{tabular}{ll|ccc|c|c}
        \textbf{\textit{Syntactic}} && \textbf{Dataset} & \textbf{Typology-Vector}& \multicolumn{3}{c}{\textbf{NDCG@5}}\\ 
        \textbf{Feature-Src} && \textbf{Features}& \textbf{Representation}& \textit{biLSTMs} & \textit{XLM-R}  &M-BERT \\ \hline
        \multirow{4}{*}{\textbf{URIEL}} &\textbf{a}& \checkmark & distance& \textbf{0.799}& 0.755 &0.654 \\ 
         &\textbf{b}&  -& distance& 0.385 & 0.643   &0.625 \\ 
         &\textbf{c}& \checkmark &  full& 0.776& \textbf{0.782} &0.680 \\ 
         &\textbf{d}&  -&  full& 0.721& 0.670   &\textbf{0.689} \\
 \cline{2-7}& \textit{Avg}& & & 0.670& 0.713& 0.662\\
         \hline \hline
        \multirow{4}{*}{\textbf{Grambank}} &\textbf{a}& \checkmark & distance& 0.768& 0.826 &0.653 \\ 
         &\textbf{b}&  -& distance& 0.447& 0.574 &0.638 \\ 
        \textbf{}  &\textbf{c}& \checkmark &  full& \textbf{0.788}& \textbf{0.827} &0.665 \\ 
         &\textbf{d}&  -&  full& 0.721& 0.707 &\textbf{0.692} \\
 \cline{2-7}& \textit{Avg}& & & 0.681& 0.734& 0.662\\ \midrule \midrule
        \multirow{2}{*}{\textbf{Avg \textit{(std)}}}& & & & 0.676 & 0.723 &0.662 \\
 & & & & \textit{(0.153)}& \textit{(0.085)}& \textit{(0.023)}\\ 

    \end{tabular}
    \caption{Average NDCG@5 for all model configurations trained on gold rankings. Every model configuration includes \textit{genetic} and \textit{geographic} features.}
    \label{combined-table}
\end{table*}
\subsubsection{Dataset Features}

We experiment with the inclusion and exclusion of dataset dependent features to assess the impact the training corpus might have on successful cross-lingual transfer.  We control for training corpus size in our gold rankings, but we do not control for any other corpus features across source languages. Therefore, it is necessary to evaluate the relevance of features like type-token ratio and word overlap.

\subsubsection{Evaluation}

As in \citet{lin-etal-2019-choosing}, we evaluate our ranking models with leave-one-out cross-validation. For each cross-validation fold, we exclude one target language from our test set of $n$ languages, and train our ranking model using gold transfer language rankings for each $n-1$ remaining languages. We then evaluate the model's performance on the held-out language. 
We evaluate our ranking models using Normalized Distributed Cumulative Gain (NCDG)\citep{10.1145/582415.582418}.

Specifically, we use NCDG@p, a metric that considers the top-p elements, which is defined by: 
\[
  NDCG@p = \frac{DCG@p}{IDCG@p},
\]
where the Discounted Cumulative Gain (DCG) at position p is defined as
\[
  DCG@p = \sum_{i=1}^{p} \frac{2^{\gamma_i} - 1}{log_2(i+1)}.
\]
$\gamma_i$ is a relevance score corresponding to the language at position $i$ of the predicted ranking that we are evaluating. For all $i \leq p$, $\gamma_i = p - i$, where $p$ represents the number of ranked items we wish to assign relevance. We set $p = 5$, meaning that the true best transfer language has a relevance score of $\gamma = 5$.  All languages below the top-5 are assigned $\gamma = 0$. The Ideal Discounted Cumulative Gain (IDCG) is calculated the same as DCG except it is calculated over the gold-standard ranking. An NCDG@p of 1 indicates that the top-p predicted elements match the top-p gold elements exactly. We report the average NDCG@5 across all $N$ leave-one-out models.

\subsection{Analyzing Feature Importance}
To compare the most relevant features for transfer in POS tagging across architectures, we use our most full featured ranking model, incorporating dataset-dependent features, \textit{syntactic} features from Grambank, and \textbf{full} Typology-Vectors. We train three rankers, one for each architecture. During training, each feature is assigned an importance score based on the gain resulting from splits made on that feature. For a given split, we calculate gain as the reduction in squared error from the parent node to the child nodes, summed across all trees in the ranking model. We report average gain over all cross-validation folds and identify the top-5 most important features for each model.

\begin{table*}[]
\setlength{\tabcolsep}{4pt}  
\centering
\begin{tabular}{lr|lr|lr}
 \multicolumn{2}{c}{XLM-R} & \multicolumn{2}{c}{M-BERT}& \multicolumn{2}{c}{BiLISTM}\\
    \toprule
    \textbf{Feature}& \textbf{Gain}  & \textbf{Feature}&\textbf{Gain}  & \textbf{Feature}& \textbf{Gain}\\ \midrule
    \textit{genetic} & 272.95  & \textit{genetic} &283.41&\textit{word\_overlap}& 264.24 \\
    \textit{word\_overlap}& 102.82  & \textit{word\_overlap}&130.90&\textit{transfer\_ttr}& 118.17 \\
    \textit{transfer\_ttr}& 67.60  & \textit{transfer\_ttr}&42.49& \textit{genetic} & 100.78  \\
    \textit{distance\_ttr}& 25.74  & \textit{distance\_ttr}&24.67& \textit{distance\_ttr}& 12.66 \\
    \textit{GB093}& 11.96  & \textit{task\_ttr}&10.06&\textit{INV\_VOW\_10\_MORE}& 7.90 \\ \bottomrule
 \textbf{Standard Deviation}& 17.08& & 17.96& &17.42\\
\end{tabular}
\caption{Feature importance for top-5 features by model for ranker trained \emph{with} dataset features and full Grambank vectors.}
\label{tab:top-dataset}
\end{table*}

\section{Results}
\subsection{Dataset vs. Typological Features}
In Table \ref{combined-table}, we observe that regardless of  \textit{syntactic} vector source, models trained with \textbf{distance} Typology-Vector representations and \textit{without} dataset features (setting \textbf{b}) perform relatively poorly. This suggests that coarse grained information from \textbf{distance} Typology-Vector representations may not be sufficient for choosing a transfer language. 
However, when we replace \textbf{distance} Typology-Vector representations with \textbf{full}, performance increases substantially. On average, NDCG@5 jumps by 0.148 between settings \textbf{b} and \textbf{d} over all 6 architecture/feature-source pairings.
The performance gains from including dataset features are even more significant. On average, NDCG@5 jumps by 0.19 between settings \textbf{b} and \textbf{a}.

These findings suggest that both fine-grained typological features \textit{and} dataset-dependent features support more accurate transfer language ranking. Both feature sources provide meaningful signals to the ranker, but setting \textbf{c} results in the best average ranker performance, suggesting that an integrated view of transfer language choice is most effective. 

M-BERT stands out as a notable outlier, as setting \textbf{d} produces the highest-performing M-BERT rankers. It is unclear why excluding dataset features benefits transfer language ranking for M-BERT. However, it is noteworthy that M-BERT exhibits by far the lowest standard deviation in performance, suggesting its rankers are less sensitive to variations in feature configuration. We leave further analysis of this phenomenon to future work.

\subsection{Grambank vs. URIEL}
Rankers leveraging Grambank \textit{syntactic} features outperform those trained with URIEL \textit{syntactic} features in ranking biLSTMs and XLM-R on average, suggesting that the typological information captured by Grambank may be more informative for selecting a transfer language. However, M-BERT is yet again an outlier-- on average, M-BERT rankers perform equivalently regardless of \textit{syntactic} feature-source.

\subsection{Feature Importance}

We investigate feature importance within our most fully-featured  ranking model, which incorporates dataset-dependent features, \textit{syntactic} features from Grambank, and full Typology-Vectors. Though this is not always the highest performing setting, it enables us to elucidate the interplay between the dataset-dependent and typological features most clearly. We identify  the top-5 most important features for each of our models in Table \ref{tab:top-dataset}.
Four out of five features are shared across architectures: genetic, word\_overlap, transfer\_ttr, and distance\_ttr. 
Notably, these are primarily dataset-dependent features. This consistency in relative feature importance across models suggests that the features that determine a suitable transfer language choice may not be architecture-dependent. 
On the other hand, it is interesting that \textit{genetic} is most important for XLM-R  and M-BERT but not for biLSTMs. It is possible that the shared representation space built during multilingual pretraining already contains features like word-overlap making them less relevant for selecting a finetuning dataset. 

\begin{table}[t]
   \fontsize{8.5}{10} \selectfont
    \centering
    \setlength{\tabcolsep}{2pt}
    \renewcommand{\arraystretch}{1.1}
    \begin{tabularx}{\linewidth}{@{}Xc@{}}
        \toprule
        \textbf{Feature} & \textbf{Gain} \\ \midrule
        \multicolumn{2}{c}{\textbf{XLM-R}} \\ 
        \textit{genetic} & 362.93 \\
        \textit{GB020} & 11.62 \\
        \textit{GB080} & 8.90 \\
        \textit{GB093} & 7.68 \\
        \textit{INV\_OPEN\_FRONT\_UNROUNDED\_VOWEL} & 7.48 \\ \hline
        \textbf{Standard Deviation} & 20.93 \\ 
        \midrule
        \midrule
        \multicolumn{2}{c}{\textbf{M-BERT}} \\ 
        \textit{genetic} & 407.08 \\
        \textit{GB022} & 8.44 \\
        \textit{GB093} & 7.07 \\
        \textit{INV\_PALATAL\_LATERAL\_APPROXIMANT} & 6.42 \\
        \textit{GB020} & 6.39 \\
        \textit{GB114} & 5.32 \\  \hline
        \textbf{Standard Deviation} & 23.46 \\
        \midrule
        \midrule
        \multicolumn{2}{c}{\textbf{biLSTM}} \\ 
        \textit{genetic} & 342.61 \\
        \textit{INV\_OPEN\_MID\_CENTRAL\_UNROUNDED\_VOWEL} & 21.75 \\
        \textit{GB172} & 19.12 \\
        \textit{INV\_MID\_CENTRAL\_UNROUNDED\_VOWEL} & 17.66 \\
        \textit{INV\_LABIODENTAL\_NASAL} & 12.22 \\ \hline
        \textbf{Standard Deviation} & 19.83 \\ 
        \bottomrule
    \end{tabularx}
    \caption{Feature importance for rankers Trained with full Grambank vectors and \emph{without} dataset features}
    \label{tab:top-feats-no-dataset}
\end{table}

\begin{table*}[h!]
    \small
    \centering
    \begin{tabular}{llll|llll}
        \toprule
        \textbf{Src/Tgt} & \textbf{XLM-R Rank} & \textbf{BiLSTM Rank} & \textbf{Diff.} & \textbf{Src/Tgt} & \textbf{XLM-R Rank} & \textbf{BiLSTM Rank} & \textbf{Diff.} \\
        \midrule
        eus/cat & 354 & 22 & 332 & ukr/pol & 10 & 339 & 329 \\
        kor/cat & 360 & 29 & 331 & ces/pol & 8 & 302 & 294 \\
        kor/glg & 339 & 13 & 326 & rus/pol & 32 & 324 & 292 \\
        kor/fra & 359 & 54 & 305 & dan/fin & 66 & 345 & 279 \\
        pol/cat & 323 & 24 & 299 & rus/lav & 26 & 304 & 278 \\
        eus/glg & 301 & 12 & 289 & lav/pol & 86 & 337 & 251 \\
        eus/fra & 334 & 46 & 288 & eng/fin & 96 & 347 & 251 \\
        pol/fra & 331 & 49 & 282 & ces/rus & 15 & 258 & 243 \\
        tur/cat & 305 & 27 & 278 & ukr/lav & 56 & 297 & 241 \\
        pol/glg & 282 & 7 & 275 & fra/fin & 108 & 348 & 240 \\
        \bottomrule
    \end{tabular}
    \caption{Greatest difference in relative performance differences between XLM-R and biLSTM.  Better biLSTM performance (left) vs. better XLM-R performance (right).}
    \label{tab:rankings-comparison}
\end{table*}

\begin{table*}[h!]
    \centering
    \begin{tabular}{ll|ll}
 \multicolumn{2}{c}{XLM-R}& \multicolumn{2}{c}{biLSTM}\\
        \toprule
        \textbf{Language Family Pair} & \textbf{Count} & \textbf{Language Family Pair} & \textbf{Count} \\
        \midrule
        Indo-European/Indo-European & 125 & Basque/Indo-European & 13 \\
        Indo-European/Uralic & 14 & Koreanic/Indo-European & 14 \\
        Austronesian/Indo-European & 5 & Indo-European/Indo-European & 71 \\
        Basque/Uralic & 1 & Turkic/Indo-European & 12 \\
        Turkic/Uralic & 1 & Koreanic/Uralic & 1 \\
        Austronesian/Uralic & 1 & Koreanic/Austronesian & 1 \\
        Indo-European/Turkic & 14 & Indo-European/Uralic & 14 \\
        Indo-European/Basque & 6 & Basque/Uralic & 1 \\
        Turkic/Indo-European & 3 & Indo-European/Koreanic & 14 \\
        Basque/Indo-European & 2 & Indo-European/Austronesian & 14 \\
        Koreanic/Uralic & 1 & Turkic/Austronesian & 1 \\
        Austronesian/Turkic & 1 & Turkic/Koreanic & 1 \\
        Basque/Turkic & 1 & Basque/Austronesian & 1 \\
        Koreanic/Indo-European & 1 & Austronesian/Uralic & 1 \\
        Koreanic/Turkic & 1 & Austronesian/Indo-European & 10 \\
         &  & Austronesian/Koreanic & 1 \\
         &  & Basque/Koreanic & 1 \\
         &  & Koreanic/Basque & 1 \\
         &  & Turkic/Basque & 1 \\
         &  & Indo-European/Basque & 8 \\
         &  & Austronesian/Basque & 1 \\
         &  & Turkic/Uralic & 1 \\
        \bottomrule
    \end{tabular}
    \caption{Distribution of  language family pairs that ranked relatively higher in XLM-R performance rankings (left) vs.  those that ranked relatively higher in biLSTM performance rankings (right)}
\label{ref:language-families}
\end{table*}

\section{Supplementary Analyses}

\subsection{Excluding Dataset Features}
For the sake of comparison, we also analyze the top-5 features for a ranking model trained with \textit{syntactic} features from Grambank and \textbf{full} Typology-Vectors \textit{without} dataset-dependent features. These rankers do not consistently underperform their dataset-dependent counterparts, raising the question of which dataset-independent features carry the most weight. 

Looking at Table \ref{tab:top-feats-no-dataset}, we find that the \textit{genetic} feature yields substantially more gain  than any other feature. It is possible that \textit{genetic} scores so highly because it serves as a proxy for many of the other features. This intuition is supported by \citet{Skirgård_Haynie_Blasi_Hammarström_Collins_Latarche_Lesage_Weber_Witzlack-Makarevich_Passmore_et_al._2023}, who show that phylogenetic relationships explain a majority of the variance in all but a few Grambank features.

Other than \textit{genetic}, M-BERT and XLM-R seem to share more top features with each other than with biLSTMs-- GB093 and GB020 both ranking highly. However, this does not necessarily indicate a meaningful difference between the architectures. Excluding \textit{genetic}, gain is relatively low and consistent across features. 
This finding suggests that it may not be possible to identify especially salient fine-grained features, because relevance is distributed over the full feature set. In a sense, the whole may be greater than the sum of its parts. 

\subsection{Ranking Analysis: BiLSTMs vs. XLM-R}

To contextualize our findings, we conducted a comparative analysis of gold transfer language rankings for biLSTMs and XLM-R. For each architecture, we generated an ordered list of source-target pairs based on performance. We then compared rank differences across architectures for each pair. Table \ref{tab:rankings-comparison} highlights the top-10 language pairs with the most divergent rankings.

XLM-R performs best on language pairs within the same family or subfamily, such as Slavic pairs, likely due to better typological alignment. Meanwhile, biLSTMs excel on pairs with weaker genetic ties. To further explore these trends, we counted occurrences of language family pairs where either XLM-R or biLSTM had a relative ranking advantage in Table \ref{ref:language-families}.

We see that XLM-R comparatively excels on Indo-European/Indo-European pairs, while biLSTMs perform relatively better on unrelated or weakly related pairs. These results align with expectations: XLM-R’s zero-shot approach benefits from well-matched transfer pairs, whereas biLSTMs can make effective use of small amounts of target language training data.

\section{Conclusion}

We find that features such as word overlap, type-token ratio, and genealogical distance are consistently influential in transfer language selection regardless of model architecture; their importance may be somewhat model-agnostic.

Our findings also highlight the crucial role of dataset-dependent features in ranking transfer languages for cross-lingual transfer. Rankers trained with these features outperform those relying solely on coarse-grained typological features.

At the same time, while coarse-grained typological features alone are insufficient, rankers trained with \emph{fine-grained} typological features achieve impressive results even without dataset-dependent features. The most successful ranking performance comes from combining both dataset-dependent and fine-grained typological features, underscoring the value of a comprehensive approach to transfer language selection.

Crucially, these insights enable us to better support languages that are not well-represented in MLM pretraining. By identifying effective transfer languages with interpretable features, we can improve cross-lingual transfer for lower-resource languages, expanding the reach of NLP beyond those languages that benefit from large-scale pretraining.

\section*{Limitations}
Since the scope of this paper is limited to crosslingual transfer for POS tagging, it would be interesting to explore  whether our results are extensible to other tasks. We are also limited in that we consider a set of just 20 target languages, 13 of which are Indo-European. This paper represents a step forward
in explaining the dynamics at play in successful crosslingual transfer, but more work is necessary to determine whether our findings generalize across diverse linguistic contexts. 

\section*{Acknowledgments}
Parts of this work were supported by the National Science
Foundation under Grant No. 2149404, “CAREER:
From One Language to Another.” Any opinions,
findings, and conclusions or recommendations ex-
pressed in this material are those of the authors
and do not necessarily reflect the views of the Na-
tional Science Foundation. This work utilized the
Blanca condo computing resource at the University
of Colorado Boulder. Blanca is jointly funded by
computing users and the University of Colorado
Boulder.

\bibliography{custom}
\appendix

\end{document}

%% file: alt_intro.tex
Despite being trained on 100+ languages, pretrained multilingual language models (MLMs) fail to cover the vast majority of the world's languages. Finetuning MLMs for zero-shot cross-lingual transfer is a useful technique to extend their reach by circumventing the lack of task-specific labeled data in low-resource languages. Effective zero-shot transfer hinges on choosing an appropriate source language \cite{Eronen_Ptaszynski_Masui_2023, Eronen_Ptaszynski_Masui_Arata_Leliwa_Wroczynski_2022, layacan-etal-2024-zero}, but it is still not well understood how to make this selection. Most analyses of successful source/target pairs fall into one of two categories: typological or dataset-dependent. The typological view investigates the role of linguistic similarity, with studies showing that more "similar" languages tend to form better source/target pairs \cite{Eronen_Ptaszynski_Masui_2023, de-vries-etal-2022-make, Lauscher_Ravishankar_Vulić_Glavaš_2020}. Much of this typological analysis is coarse-grained, focusing on features like language family or abstract distance measures. The dataset-dependent view focuses on comparing source and target datasets based on features like sub-word overlap \cite{wu-dredze-2019-beto, pires-etal-2019-multilingual,K2020Cross-Lingual}. Few papers consider both views, and those that do focus on older methods of crosslingual transfer like bilingual LSTMS \cite{lin-etal-2019-choosing}. Additionally, previous analyses shed little light on the linguistic question of which fine-grained typological features are especially relevant for the task.

This primary goal of this paper is to offer a deeper understanding of effective transfer language selection across architectures, comparing  crosslingual transfer with biLSTMs to XLM-R \cite{conneau-etal-2020-unsupervised} and M-BERT \cite{devlin-etal-2019-bert}. We aim to identify which features contribute to selecting a successful source/target pair for part-of-speech (POS) tagging. We focus on POS tagging because it directly reflects typological features such as word order. Our analysis addresses the following key questions: 
\begin{itemize}
    \item[\textbf{Q1.}] Which features are most important for cross-lingual transfer? 
    \item[\textbf{Q2.}]Do these features differ between biLSTMs and MLMs?
    \item[\textbf{Q3.}] How does the granularity of typological features—whether fine or coarse—affect transfer language selection? 
    \item[\textbf{Q4.}] Is it necessary to consider data set features in selecting a transfer language? 
\end{itemize}  

We train a series of gradient-boosted decision tree models to rank transfer languages for POS tagging, with separate rankers for the two architectures. 
During training, we generate feature importance scores and identify the most salient features for each architecture (Q1, Q2). To examine the role of fine-grained typological features, we compare two typological inputs: source/target distance measures, and full finegrained feature vectors (Q3). We also evaluate how the source and quality of typological data affects ranker performance by swapping between URIEL \cite{littell-etal-2017-uriel} and Grambank \cite{Skirgård_Haynie_Blasi_Hammarström_Collins_Latarche_Lesage_Weber_Witzlack-Makarevich_Passmore_et_al._2023} feature vectors. Last, we investigate whether typological information alone can effectively determine suitable source/target language pairs by experimenting with the exclusion of dataset-specific features (Q4). 

We find that impressive performance can be achieved when relying primarily on either feature category, without the need for the other, indicating that both "typological" and "dataset-dependent" views of transfer language choice represent independently viable strategies. However, peak performance is achieved by combining dataset-dependent and fine-grained typological features.
Crucially, our analysis reveals that key features such as word overlap, type-token ratio, and genealogical distance remain consistently important across architectures, suggesting that the relevance of these features may transcend specific model designs, offering broader insights into cross-lingual transfer that could enable us to better leverage MLMs for low-resource applications.